# On the Impact of Class Imbalance on the Learning Dynamics of Deep Neural Networks: An Intuitive Insight

Ismail B. Mustapha
Faculty of Computing
Universiti Teknologi Malaysia
Johor, Malaysia
bmismail2@live.utm.my

Shafaatunnur Hasan
Faculty of Computing
Universiti Teknologi Malaysia
Johor, Malaysia
shafaatunnur@utm.my

Sunday O Olatunji
Faculty of Computing
Adejkunle Ajasin University
Akungba-Akoko, Nigeria
oluolatunji.aadam@gmail.com

Hatem S Y Nabus
Faculty of Computing
Universiti Teknologi Malaysia
Johor, Malaysia
hatemrun@ gmail.com

***Abstract*—Class imbalance in deep neural networks (DNNs) has witnessed a rapid increase in research attention in recent years. However, the varying accounts of the reasons behind the poor performance of DNN on imbalance data in pertinent literature shows that little is known about how this agelong phenomenon impacts the performance of DNNs. A better understanding of this problem is crucial to developing effective DNN-based imbalance methods. Thus, this study systematically investigates the impact of class imbalance on the learning dynamics of DNN by monitoring the learning pattern of DNN models on both the majority and minority classes of datasets of varying imbalance ratios. Experimental findings shows that as against learning from balanced datasets where DNN learns the classes similarly, class imbalance has severe deteriorating impact on the performance of DNN, driving the model to underfit the minority class samples in the early training epochs while simultaneously learning only the majority class. Although DNN ultimately learns the minority samples, learning in this manner only results in learnt minority representations that are non-generalizable at test phase because they are merely overfitted to keep the overall training loss as low as possible.**

## I. Introduction

One of the most prominent supervised machine learning (ML) challenges of the last couple of decades is the class imbalance or imbalanced data problem [1]. This occurs when the distribution of samples across categories or classes of a dataset is substantially skewed such that most of the data samples belong to a subset of the classes, leaving the other classes with a few. Given a typical case of a binary classification problem, if most of the data instances belong to one of the classes (majority class) while the few remaining samples belong to the other (minority class), the data is said to be imbalanced and learning from such is described as learning from imbalance data [2]. This problem naturally occurs in many real-life application domains of ML such as flight delay prediction [3, 4], fraud detection [5-7], chronic disease diagnosis [8, 9], oil spillage detection in radar images amongst many others.

In the context of classical ML, concerted efforts have been made over the past years to understand the reason behind the poor performance of ML algorithms on the minority class instances of imbalanced data. Relevant studies in this regard have often attributed the poor performance of classical ML methods to underfitting of the minority classes during the training phase, with the main reason being the lack of sufficient data or requisite information needed to learn generalizable decision boundaries of the minority class samples [10-12]. Also, other studies such as [1, 13], have added that the degree of separability or overlapping of the classes, also known as data complexity [14], remains a major cause of the poor performance of ML algorithms on imbalance data, citing that linearly separable data are unaffected by class imbalance. However, real-world data are rarely linearly separable, given the intricate interactions of systems from which the data are collected.

A class of ML methods the has in recent years gained widespread adoption is the Deep neural networks (DNN). DNN are a class of artificial neural networks-based ML approaches characterized with multiple hidden layers and units used to progressively learn high level representations or features of a given input for improved performance [15, 16]. DNN methods offer additional advantages over shallow NN and traditional ML methods in that they are suitable for large datasets and multimodal learning problems which makes them more adaptable to real-world problems [17]. Thus, DNN models have continued to witness tremendous success in numerous ML application domains in recent times; particularly in computer vision, natural language processing (NLP), audio/speech recognition and machine translation tasks where they continue to advance the state-of-the-art [18]. Despite these remarkable success, pertinent studies have shown that DNN are not immune to the class imbalance problem [19, 20]. DNN models have been found to perform poorly on the minority classes when dealing with imbalance data [19, 21-23].

Several studies have in recent years sought to explain the reason behind the poor performance of DNN models on the minority samples when learning from imbalance data, a few of which have identified underfitting of the minority class as the major reason [24, 25]. The proponents of this view have suspected underfitting of the less represented classes as the reason for the suboptimal model performance mainly because of the smaller size of the minority class. On the other hand, a sizeable number of representative studies have attributed the class imbalance problem to overfitting of the minority class samples [21, 22, 26, 27]. While those who hold this view believe that unlike classical ML methods that underfit the minority class, the overparametrized nature of DNN empowers it to fit the training data to near-zero loss, they differ in why the resulting model is unable to generalize to the minority class at test phase. Some claim the resulting model is unable to generalize the learnt representations of minority class to the test data due to feature deviation (i.e., the learnt training features deviates from that of the test data with level of deviation increasing with decreasing minority class samples) [22, 26], whereas some other studies argue that it is due to biased decision boundary, where because the decision boundary the model learns is close to minority features with little or no margin for error, the DNN model easily classifies the test minority samples as the majority class which has a larger margin between its features and the decision boundary [21, 27-29]. In addition, the disparity in the training and test data distribution has also been hinted in [30] as a possible reason for poor performance of DNN models on the minority class samples. While this can manifest in different ways such as concept drift, where the target variable changes in an unforeseen way over time, the test set of popular long-tailed benchmarks like CIFAR10 and CIFAR100 are characteristically balanced.

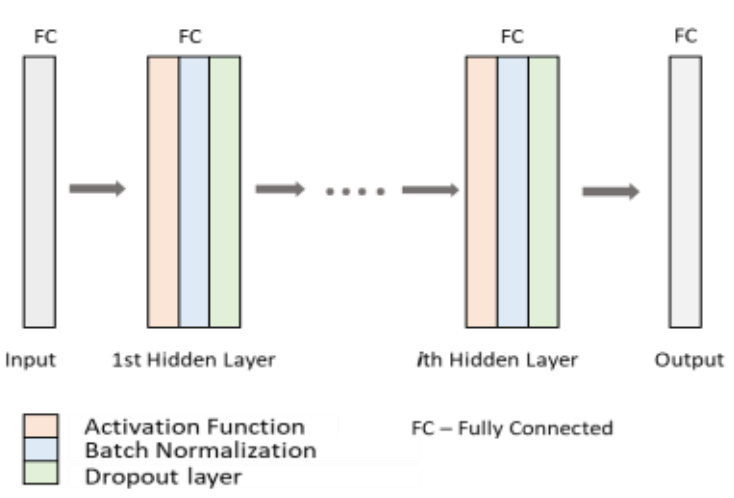

Fig. 1 DNN Architecture

Although, there has been notable growth in studies relating to approaches for addressing class imbalance in DNN in recent years, most of these studies have proposed imbalanced methods based on the general notion that class imbalance hinders optimal deep learning, with little or no understanding of how and why this occurs. Besides, the varying account of the possible reasons behind the poor performance of DNN on imbalance data from the different aforementioned studies suggest the need for a deeper understanding of this phenomenon. Thus, this research aims to systematically investigate how class imbalance affects the learning dynamics of DNN by monitoring the learning pattern of DNN models for both the majority and minority classes of datasets of varying imbalance ratios over the training epochs.

This article is organized as follows. Section 2 presents the materials and methods of the study, followed by the results and discussion in Section 3. The conclusion and future direction are presented in Section 4.

## II. Materials and Methods

### *A. Datasets*

30 binary imbalanced benchmark datasets from UCI[1] and Keel[2] data repositories have been carefully selected for this study. Details of each dataset are presented in Table A.1 of the Appendix in descending order of imbalance ratio (IR). The higher the IR, the more imbalanced the dataset is and vice versa. Also, the silhouette coefficient (SC) [31] for each dataset has been included to indicate the degree of separability of class clusters. where values close to +1 generally indicate well separated clusters and values close to 0 indicate some degree of overlap in the class clusters. Values close to -1 indicate a greater degree of overlap in which samples from a cluster are well in the cluster of another class.

Also, 3 balanced datasets were created from Iris dataset (a 3-class multiclass dataset obtained for UCI database) by pairing the classes to form setosa_vs_versicolor, setosa_vs_virginica and virginica_vs_versicolor respectively. The details of each balanced dataset are presented in Table A.2 of the Appendix. These are to serve as balanced benchmark datasets to which the imbalanced ones are compared..

### *B. Model Architecture*

The template is used to format your paper and style the text A fully connected DNN architecture based on ReLU, activation function (AF) is used for modelling each imbalanced data in this study. Batch normalization and 0.5 dropout rate are applied after each AF to avoid overfitting. Fig. 1 shows the structure of each DNN architecture. FC means each layer is made of fully connected neurons.

### *C. Hyperparameter Optimization and Model Training*

Hyperparameter optimization of each model was done using 80% of each dataset. Similar to [32], only the depth and width of each model are optimized. For each ReLU-based model, a network width of 50 neurons per hidden layer was found to be sufficient for the models to overfit the data after experimenting with widths of 512, 300, 100, 50 and 32 neurons respectively. The depth of these models was optimized starting with a depth of two (i.e., 2-hidden layers) and varying it up to six, optimal DNN architectures for each datasets was determined via the best mean AUC over 5-fold cross validation as in [33].

Each of the optimized models per dataset described above were trained and tested using 80% and 20% split of each dataset respectively. The model weights were randomly initialized with uniform distribution and Xavier variance [34] with zero bias before training with stochastic gradient descent minibatch sizes ranging from 1/32 to 1/100 of the respective datasets sample sizes. Adam stochastic optimizer [35] with learning rate of 0.001 was used for model optimization and cross entropy loss as loss function. Each model was trained to stop after a minimum of 250 epochs.

### *D. Evaluation Metrics*

Given that the aim of this study is to investigate how class imbalance impacts optimal learning in DNN, special attention is paid to the accuracy of the model on the respective classes (majority and minority classes). Hence, in addition to the model loss, the training and test accuracies of the minority class (True Positive Rate (TPR)) and the majority class (True Negative Rate (TNR)) are respectively used to evaluate the model performance. The mathematical formula for each measure is given by equations (1) and (2), where $TP$, $FN$, $TN$ and $FP$ stand for true positive, false negative, true negative and false positive counts respectively.

$$TPR = \frac{TP}{TP+FN} \quad (1)$$

$$TNR = \frac{TN}{TN+FP} \quad (2)$$

[1]https://archive.ics.uci.edu/ml/datasets.php

[2] https://sci2s.ugr.es/keel/index.php

## III. Results and Discussion

To understand how class imbalance affects the classification performance of DNN, the overall model training loss as well as the respective training losses of the minority and majority classes over the entire training epochs of each DNN model were studied alongside the training accuracy of the model on the minority (true positive rate) and majority (true negative rate) classes. To fully grasp the impact of class imbalance on the learning dynamics of DNN models, it is important to know the ideal model performance on balanced datasets. Thus, the performance of the DNN model on the 3 balanced binary datasets created from the Iris dataset as described above were studied by observing changes in the overall training loss and the respective per-class training losses and accuracies over the training epochs.

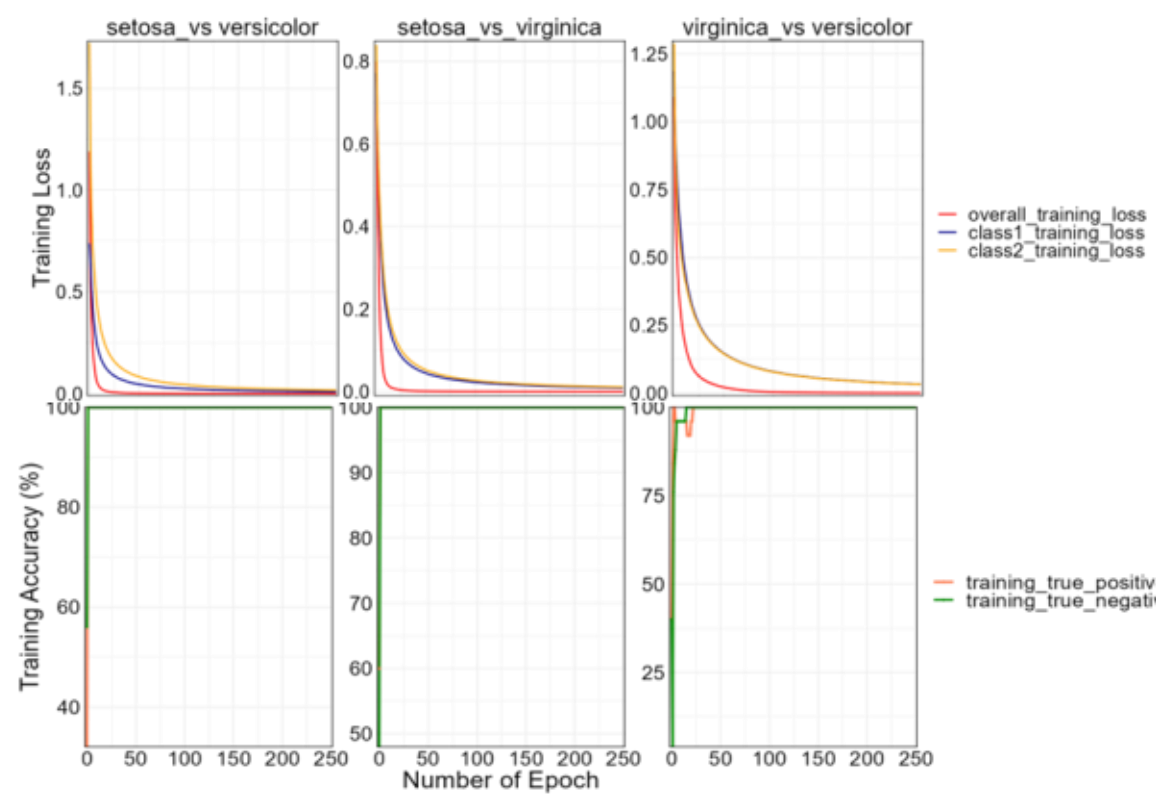

Fig. 2 Variation in Training Losses and Accuracies of Balanced Datasets Across Training Epochs

Although loss in ML and specifically DL is an indication of how well a model is learning a given task, it can vary considerably across datasets. Hence, the per class accuracy (true positive (TP) and true negative (TN) rates) of the model for each dataset is also included to serve as a quantitative measure of the performance of the model on each class as training progresses. The plot of the training losses (above) and the respective per-class training accuracies (below) for each balanced dataset over the training period are presented in Fig. 2. Thus, changes in loss values and per class accuracies over the training epochs are used to illustrate the performance of the model on each dataset. In terms of the training losses, it can be observed that the loss of each class decreases at roughly the same rate across the datasets; steadily descending at roughly the same pace before plateauing as they approach zero. This is further indicated by the per class training accuracies of the model across the datasets where 100% is achieved for each class in the first few training epochs. Understandably, virginica_vs_versicolor is the most complex of the balanced datasets given a lower SC of 0.3316. Nevertheless, the model learns both classes equally and achieved 100% training accuracies on both classes within the first 20 epochs. Additionally, Table A.2 gives a quantitative perspective of the training and test (generalization) performance of the DNN model on each balanced dataset. As can be observed, the model generalizes equally to the respective classes of the test set of each imbalanced dataset.

The overall training loss as well as the per class training losses and accuracies for the majority and minority classes of each binary imbalanced dataset can be found in Fig A.1. The datasets are respectively prefixed with 1 to 30 to indicate their position in descending order of IR i.e., 1.abalone19 and 30.pima are the most and least imbalanced datasets respectively. In sharp contrast to what was observed in the balanced scenario in Fig. 2, the plots in Fig A.1 show interesting trends that provide intuitive understanding of the characteristic poor performance of DNNs on minority samples when learning from imbalanced data. In relation to the majority class, it can be observed that the training loss of the majority class exhibits a similar pattern as the overall training loss for each dataset; decreasing at roughly the same rate before similarly plateauing as training progresses. The training accuracies of the respective majority classes across the imbalanced datasets as indicated by the true negative rates also reflects this pattern with the model reaching at least a true negative rate of 99% within the first 3 training epochs across the datasets.

TABLE I. Training and Test Performance on Balanced Datasets

| Dataset | Training Phase | | | Test Phase | | |
|---|---|---|---|---|---|---|
| | *TN Rate* | *TP Rate* | *Total Accuracy* | *TN Rate* | *TP Rate* | *Total Accuracy* |
| setosa_vs vesicolor | 100.0 | 100.0 | 100.0 | 100.0 | 100.0 | 100.0 |
| setosa_vs_virginica | 100.0 | 100.0 | 100.0 | 100.0 | 100.0 | 100.0 |
| virginica_vs_versicolor | 100.0 | 100.0 | 100.0 | 90.0 | 100.0 | 95.0 |

On the other hand, the respective minority class losses of the imbalance datasets show a considerably different trend. The minority class losses mostly increase in the early training epochs before decreasing. During the early training epochs, the training accuracy of the minority class of each imbalance dataset, as indicated by the true positive rates reduces to 0% before increasing afterwards, indicating that the DNN models initially learns the majority class during training while neglecting the minority class samples i.e., all the training data irrespective of the class they belong are mostly classified as the majority class in the early training epochs of the DNN model after which it begins to distinguish the minority class samples. In other words, the minority samples are essentially underfitted earlier in the training process. Learning one class while neglecting the other obviously defies equity considerations and fairness across groups that should be in place during training [36]. Despite the early episodes of underfitting, the overparameterized nature of DNNs allows them to interpolate training samples and their respective labels over time [37]. Thus, the model eventually learns the minority class samples after the early underfitting phase, though only after the majority ones have been learnt. As shown in Fig 3, the learnt minority representations are not generalizable given a generally lower true positive rate on the test set compared to the respective true negative rates and the overall accuracies. This also shows that the generally high test accuracies mainly reflect the model performance on the majority samples. Overall, despite training true positive and true negative rates of 100% respectively, the DNN models were unable to generalize learnt minority representations at test phase, suggesting that the minority samples were overfitted to keep the training loss as minimum as possible. While this trend is observable across the various datasets, the severity can be seen to generally decrease with decreasing IR, as reflected by the reduced margins between the true positive and true negative rates as IR decreases.

Nonetheless, there are few exceptions amongst the datasets where impact of class imbalance is minimal. For instance, the model learns the majority and minority class samples of the 3.kddcup-rootkit-imap_vs_back dataset (IR=110.25) equally as shown by its training losses and true negative and positive rates in Fig A.1 and generalizes the learnt representations for both classes well at test phase. Although this dataset is the third most imbalanced binary benchmark in this study, a silhouette coefficient (SC) of 0.91 shows that the class clusters of this dataset are well separated. Thus, the classes of 3.kddcup-rootkit-imap_vs_back are learnt at the same rate without preference for one over the other and do not suffer the early underfitting episode suffered by other benchmarks with much lower IRs but some degree of overlap. Indicating that the generally held notion that class imbalance is not an issue when class clusters are well separated also holds in DNN [38].

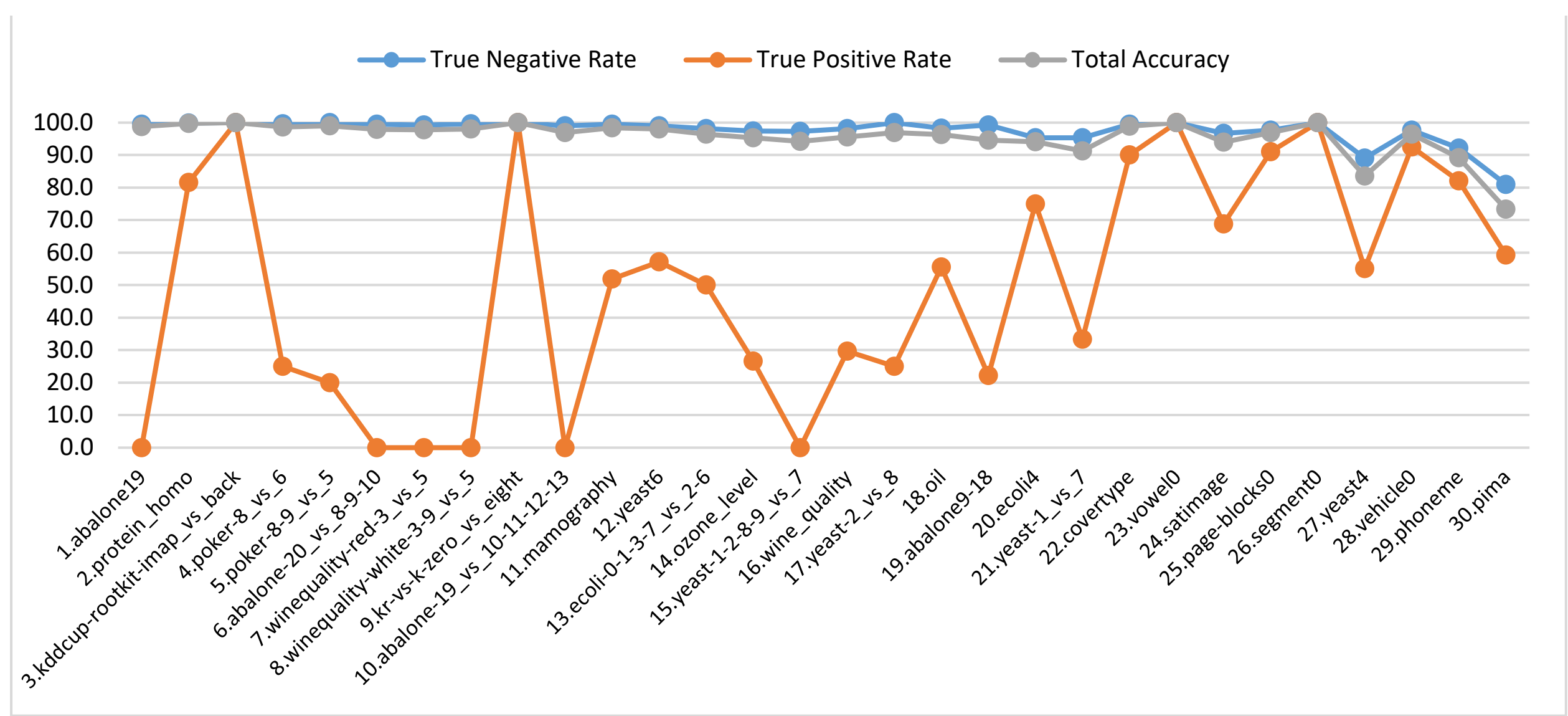

Fig. 3 Test Performance on Imbalanced Datasets

Also, the size of the minority samples is shown to play an important role when learning from an imbalance dataset. For instance, the highly imbalanced 2.protein_homo benchmark (IR=111.46) with a lesser degree of complexity (given SC of 0.56) than 3.kddcup-rootkit-imap_vs_back has over 1290 minority samples which holds useful information to ensure the early underfitting episodes is avoided. However, the DNN model clearly learns the minority class samples at a slower pace compared to the majority ones as shown by its per class training losses and accuracies. Similar trend is also observable on the 11.mamography (IR=42.01, SC=0.45) and 22.covertype (IR=13.02, SC=0.11) datasets. Compared to 13.ecoli-0-1-3-7_vs_2-6 (IR=36.33, SC=0.46) benchmarks with similar degree of complexity as 11.mamography, the former can be seen to suffer early underfitting episodes whereas the latter does not despite the latter having a higher IR. The likely explanation for this is that the class imbalance problem is further exacerbated by the meagre sample size of 13.ecoli-0-1-3-7_vs_2-6 (281 samples only 7 of which are minority samples). Thus, for datasets with low IR such as 23.vowel0 and below, the early underfitting episodes are notably minimal, although the model learns the majority class with higher preference. Overall, the DNN model can fit the training samples to near perfection, albeit later in the training process.

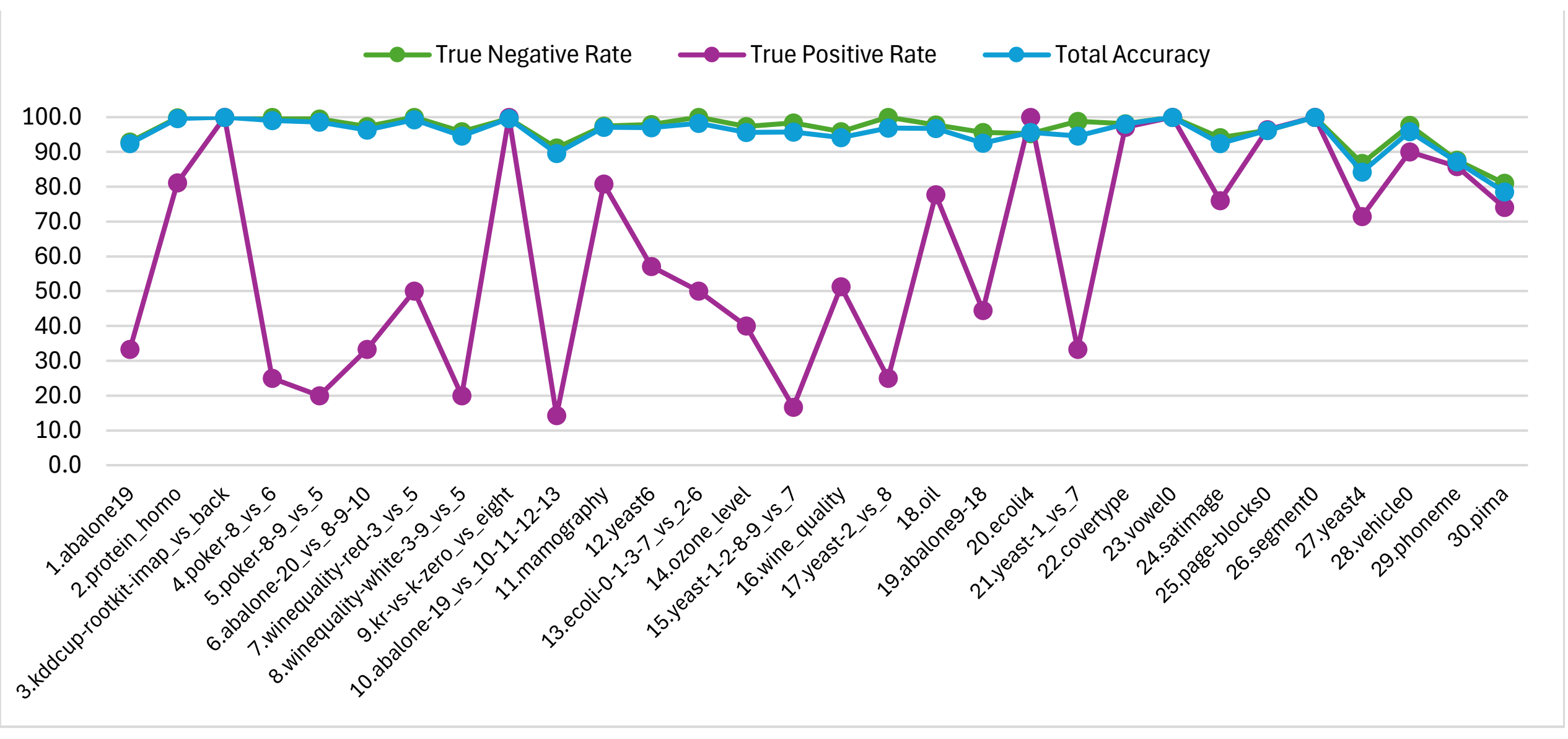

Fig. 4 Test Performance on Randomly Oversampled Imbalanced Datasets

While this line of investigation has rarely been explored in the past, there have been suggestions in pertinent works that samples from the minority class tend to have higher losses compared to samples of the majority class given the less informative nature of the learned minority class features [39]. The findings here put this suggestion into context; leveraging this knowledge to expose how this affects DNN learning from imbalance data. A corroborative finding has also been reported in [22], where the gradients of the majority and minority class samples were examined to arrive at similar findings

Additionally, random oversampling, a canonical baseline imbalance method for benchmarking imbalance methods in DNN is used to address imbalance in each dataset before training. Each DNN model was trained to stop early after 10 epochs if there are no improvement in accuracy, using the same configuration described in Section II-C. The test performance in terms of true positive rate, true negative rate as well as the overall accuracy are shown in Fig 4. It can be observed that while there is a general improvement in the true positive rate across the datasets compared to Fig 3, there remains a large margin for improvement, particularly for the high imbalance datasets. Besides, the marginal improvement in recognition of the minority class samples is mostly at the expense of the majority ones.

Also, while the presented results are regarding binary imbalanced data, the negative effect of imbalance can be more severe in the multiclass classification problems, given that multiclass classification is a generalization and more complex case of binary classification. A key observation from the foregoing is that class imbalance has a severe deteriorating effect on the classification of DNNs and worsens with increasing imbalance ratio. Other factors such as data complexity and availability when combined with class imbalance further complicate learning.

## IV. Conclusion

Conclusively, this study intuitively investigates the impact of class imbalance on the classification performance of DNN models across imbalanced datasets of varying imbalance ratios. By systematically studying the performance of DNN models on the majority and minority class samples, the following conclusions are arrived at:

- DNN can fit the datasets (balanced and imbalanced) to near-zero loss given enough training time.
- Class imbalance has deteriorating effect on the classification performance of DNNs and the severity of the effect increases with increasing imbalance ratio as well as the complexity of the dataset.
- Unlike balanced datasets where DNNs learn both majority and minority classes with roughly equal preference given similar loss values of the constituent classes, DNNs learn imbalance datasets by initially fitting the majority class samples before attempting same for the minority samples which mostly come at the cost of generalization of a DNN to the minority class samples during the test/validation phase. Thus, DNNs learn the majority class with higher preference given its lower loss values compared to the minority class.
- The trajectory of the majority class loss is mostly similar to the overall training loss; indicating that the overall training loss of imbalance data is mostly influenced by the dominant class.
- DNNs are not affected by class imbalance when learning from imbalanced datasets with well separated class clusters (i.e. no class overlaps).
- Applying random oversampling leads to some improvement in minority class recognition which varies as IR increases. This comes at some cost of recognition of the majority class samples.

## Acknowledgment

The authors thank UTM Research Management Center for financing this research through research grant number PY/2024/00260/-Q.J130000.21A6.00P67. The authors also acknowledge the contribution of late Prof Siti Maryam Shamsuddin.

## References

[1] B. Krawczyk, "Learning from imbalanced data: open challenges and future directions," Progress in Artificial Intelligence, vol. 5, no. 4, pp. 221-232, 2016.

[2] F. Provost, "Machine learning from imbalanced data sets 101," in Proceedings of the AAAI'2000 workshop on imbalanced data sets, 2000, vol. 68, no. 2000: AAAI Press, pp. 1-3.

[3] S. Choi, Y. J. Kim, S. Briceno, D. Mavris, and Ieee, "Cost-sensitive Prediction of Airline Delays Using Machine Learning," in 2017 Ieee/Aiaa 36th Digital Avionics Systems Conference, (IEEE-AIAA Digital Avionics Systems Conference, 2017.

[4] I. B. Mustapha, S. M. Shamsuddin, and S. Hasan, "A Preliminary Study on Learning Challenges in Machine Learning-based Flight Delay Prediction," International Journal of Innovative Computing, vol. 9, no. 1, 2019.

[5] S. O. Moepya, S. S. Akhoury, and F. V. Nelwamondo, "Applying Cost-Sensitive Classification for Financial Fraud Detection under High Class-Imbalance," in 2014 IEEE International Conference on Data Mining Workshop, 14-14 Dec. 2014 2014, pp. 183-192, doi: 10.1109/ICDMW.2014.141.

[6] A. Thennakoon, C. Bhagyani, S. Premadasa, S. Mihiranga, and N. Kuruwitaarachchi, "Real-time Credit Card Fraud Detection Using Machine Learning," in 2019 9th International Conference on Cloud Computing, Data Science & Engineering (Confluence), 10-11 Jan. 2019 2019, pp. 488-493, doi: 10.1109/CONFLUENCE.2019.8776942.

[7] U. Fiore, A. De Santis, F. Perla, P. Zanetti, and F. Palmieri, "Using generative adversarial networks for improving classification effectiveness in credit card fraud detection," Information Sciences, vol. 479, pp. 448-455, 2019/04/01/ 2019, doi: https://doi.org/10.1016/j.ins.2017.12.030.

[8] C. Arizmendi, D. A. Sierra, A. Vellido, and E. Romero, "Automated classification of brain tumours from short echo time in vivo MRS data using Gaussian Decomposition and Bayesian Neural Networks," Expert Systems with Applications, vol. 41, no. 11, pp. 5296-5307, 2014/09/01/ 2014, doi: https://doi.org/10.1016/j.eswa.2014.02.031.

[9] S. Afzal et al., "A Data Augmentation-Based Framework to Handle Class Imbalance Problem for Alzheimer's Stage Detection," IEEE Access, vol. 7, pp. 115528-115539, 2019, doi: 10.1109/ACCESS.2019.2932786.

[10] C. Seiffert, T. M. Khoshgoftaar, J. Van Hulse, and A. Napolitano, "Mining data with rare events: a case study," in 19th IEEE International Conference on Tools with Artificial Intelligence (ICTAI 2007), 2007, vol. 2: IEEE, pp. 132-139.

[11] J. M. J. M. Khoshgoftaar, "Survey on deep learning with class imbalance," Journal of Big Data, vol. 6, no. 1, 2019, doi: 10.1186/s40537-019-0192-5.

[12] J. Li, S. Fong, S. Mohammed, J. Fiaidhi, Q. Chen, and Z. Tan, "Solving the under-fitting problem for decision tree algorithms by incremental swarm optimization in rare-event healthcare classification," Journal of Medical Imaging and Health Informatics, vol. 6, no. 4, pp. 1102-1110, 2016.

[13] N. Japkowicz, "The class imbalance problem: Significance and strategies," in Proc. of the Int'l Conf. on Artificial Intelligence, 2000, vol. 56: Citeseer, pp. 111-117.

[14] P. Vuttipittayamongkol, E. Elyan, and A. Petrovski, "On the class overlap problem in imbalanced data classification," Knowledge-based systems, vol. 212, p. 106631, 2021.

[15] L. Deng and D. Yu, "Deep learning: methods and applications," Foundations and trends in signal processing, vol. 7, no. 3–4, pp. 197-387, 2014.

[16] Y. Bengio, A. Courville, and P. Vincent, "Representation learning: A review and new perspectives," IEEE transactions on pattern analysis and machine intelligence, vol. 35, no. 8, pp. 1798-1828, 2013.

[17] V. Borisov, T. Leemann, K. Seßler, J. Haug, M. Pawelczyk, and G. Kasneci, "Deep neural networks and tabular data: A survey," arXiv preprint arXiv:2110.01889, 2021.

[18] I. Goodfellow, Y. Bengio, and A. Courville, Deep learning. MIT press, 2016.

[19] M. Buda, A. Maki, and M. A. Mazurowski, "A systematic study of the class imbalance problem in convolutional neural networks," Neural Networks, vol. 106, pp. 249-259, 2018/10/01/ 2018, doi: https://doi.org/10.1016/j.neunet.2018.07.011.

[20] T. Grósz and I. N. T., "Document Classification with Deep Rectifier Neural Networks and Probabilistic Sampling," in Text, Speech and Dialogue, 2014, doi: 10.1007/978-3-319-10816-2_14. [Online]. Available: http://link.springer.com/chapter/10.1007/978-3-319-10816-2_14

[21] K. Cao, C. Wei, A. Gaidon, N. Arechiga, and T. Ma, "Learning Imbalanced Datasets with Label-Distribution-Aware Margin Loss," arXiv preprint arXiv:1906.07413, 2019.

[22] H.-J. Ye, D.-C. Zhan, and W.-L. Chao, "Procrustean training for imbalanced deep learning," in Proceedings of the IEEE/CVF International Conference on Computer Vision, 2021, pp. 92-102.

[23] I. B. Mustapha, S. Hasan, H. S. Nabbus, M. M. A. Montaser, S. O. Olatunji, and S. M. Shamsuddin, "Investigating Group Distributionally Robust Optimization for Deep Imbalanced Learning: A Case Study of Binary Tabular Data Classification," International Journal of Advanced Computer Science and Applications, vol. 14, no. 2, 2023.

[24] A. Galdran, G. Carneiro, and M. A. González Ballester, "Balanced-mixup for highly imbalanced medical image classification," in Medical Image Computing and Computer Assisted Intervention–MICCAI 2021: 24th International Conference, Strasbourg, France, September 27–October 1, 2021, Proceedings, Part V 24, 2021: Springer, pp. 323-333.

[25] E. Wu, H. Cui, and R. E. Welsch, "Dual Autoencoders Generative Adversarial Network for Imbalanced Classification Problem," IEEE Access, vol. 8, pp. 91265-91275, 2020.

[26] H.-J. Ye, H.-Y. Chen, D.-C. Zhan, and W.-L. Chao, "Identifying and compensating for feature deviation in imbalanced deep learning," arXiv preprint arXiv:2001.01385, 2020.

[27] Z. Li, K. Kamnitsas, and B. Glocker, "Overfitting of neural nets under class imbalance: Analysis and improvements for segmentation," in International Conference on Medical Image Computing and Computer-Assisted Intervention, 2019: Springer, pp. 402-410.

[28] B. Kim and J. Kim, "Adjusting decision boundary for class imbalanced learning," IEEE Access, vol. 8, pp. 81674-81685, 2020.

[29] X. Yin, X. Yu, K. Sohn, X. Liu, and M. Chandraker, "Feature transfer learning for face recognition with under-represented data," in Proceedings of the IEEE/CVF Conference on Computer Vision and Pattern Recognition, 2019, pp. 5704-5713.

[30] J. Ren et al., "Balanced meta-softmax for long-tailed visual recognition," arXiv preprint arXiv:2007.10740, 2020.

[31] P. J. Rousseeuw, "Silhouettes: a graphical aid to the interpretation and validation of cluster analysis," Journal of computational and applied mathematics, vol. 20, pp. 53-65, 1987.

[32] J. M. Johnson and T. M. Khoshgoftaar, "Deep Learning and Data Sampling with Imbalanced Big Data," in 2019 IEEE 20th International Conference on Information Reuse and Integration for Data Science (IRI), 30 July-1 Aug. 2019 2019, pp. 175-183, doi: 10.1109/IRI.2019.00038.

[33] D. Díaz-Vico, A. R. Figueiras-Vidal, and J. R. Dorronsoro, "Deep MLPs for Imbalanced Classification," in 2018 International Joint Conference on Neural Networks (IJCNN), 8-13 July 2018 2018, pp. 1-7, doi: 10.1109/IJCNN.2018.8489504.

[34] X. Glorot and Y. Bengio, "Understanding the difficulty of training deep feedforward neural networks," in Proceedings of the thirteenth international conference on artificial intelligence and statistics, 2010: JMLR Workshop and Conference Proceedings, pp. 249-256.

[35] D. P. Kingma and J. Ba, "Adam: A method for stochastic optimization," arXiv preprint arXiv:1412.6980, 2014.

[36] M. J. Smith, R. Axler, S. Bean, F. Rudzicz, and J. Shaw, "Four equity considerations for the use of artificial intelligence in public health," Bulletin of the World Health Organization, vol. 98, no. 4, p. 290, 2020.

[37] C. Liu, L. Zhu, and M. Belkin, "Loss landscapes and optimization in over-parameterized non-linear systems and neural networks," arXiv preprint arXiv:2003.00307, 2020.

[38] A. Ali, S. M. Shamsuddin, and A. L. Ralescu, "Classification with class imbalance problem: a review," Int. J. Advance Soft Compu. Appl, vol. 7, no. 3, pp. 176-204, 2015.

[39] Y. Cui, M. Jia, T.-Y. Lin, Y. Song, and S. Belongie, "Class-balanced loss based on effective number of samples," in Proceedings of the IEEE Conference on Computer Vision and Pattern Recognition, 2019, pp. 9268-9277.

## V. APPENDIX

Table A.1 Binary Imbalanced Datasets

| **Data** | **#Instances** | **#Attributes** | **%Majority Class** | **%Minority Class** | **IR** | **SC** |
|---|---|---|---|---|---|---|
| abalone19 | 4174 | 8 | 99.23 | 0.77 | 129.44 | -0.02 |
| protein_homo | 145751 | 74 | 99.11 | 0.89 | 111.46 | 0.56 |
| kddcup-rootkit-imap_vs_back | 2225 | 41 | 99.01 | 0.99 | 100.14 | 0.91 |
| poker-8_vs_6 | 1477 | 10 | 98.85 | 1.15 | 85.88 | -0.03 |
| poker-8-9_vs_5 | 2075 | 10 | 98.80 | 1.20 | 82.00 | 0.02 |
| abalone-20_vs_8-9-10 | 1916 | 8 | 98.64 | 1.36 | 72.69 | 0.09 |
| winequality-red-3_vs_5 | 691 | 11 | 98.55 | 1.45 | 68.10 | 0.16 |
| winequality-white-3-9_vs_5 | 1482 | 11 | 98.31 | 1.69 | 58.28 | 0.27 |
| kr-vs-k-zero_vs_eight | 1460 | 6 | 98.15 | 1.85 | 53.07 | 0.04 |
| abalone-19_vs_10-11-12-13 | 1622 | 8 | 98.03 | 1.97 | 49.69 | -0.05 |
| mamography | 11183 | 6 | 97.68 | 2.32 | 42.01 | 0.45 |
| yeast6 | 1484 | 8 | 97.64 | 2.36 | 41.40 | 0.13 |
| ecoli-0-1-3-7_vs_2-6 | 281 | 7 | 97.51 | 2.49 | 39.14 | 0.46 |
| ozone_level | 2536 | 72 | 97.12 | 2.88 | 33.74 | -0.05 |
| yeast-1-2-8-9_vs_7 | 947 | 8 | 96.83 | 3.17 | 30.57 | 0.06 |
| wine_quality | 4898 | 11 | 96.26 | 3.74 | 25.77 | 0.15 |
| yeast-2_vs_8 | 482 | 8 | 95.85 | 4.15 | 23.10 | 0.39 |
| oil | 937 | 49 | 95.62 | 4.38 | 21.85 | 0.08 |
| abalone9-18 | 731 | 8 | 94.25 | 5.75 | 16.40 | 0.11 |
| ecoli4 | 336 | 7 | 94.05 | 5.95 | 15.80 | 0.23 |
| yeast-1_vs_7 | 459 | 7 | 93.46 | 6.54 | 14.30 | 0.1 |
| covertype | 38501 | 54 | 92.87 | 7.13 | 13.02 | 0.11 |
| vowel0 | 988 | 13 | 90.89 | 9.11 | 9.98 | 0.17 |
| satimage | 6435 | 36 | 90.27 | 9.73 | 9.28 | -0.13 |
| page-blocks0 | 5472 | 10 | 89.78 | 10.22 | 8.79 | 0.51 |
| segment0 | 2308 | 19 | 85.75 | 14.25 | 6.02 | -0.06 |
| yeast4 | 1484 | 8 | 83.56 | 16.44 | 5.08 | 0.39 |
| vehicle0 | 846 | 18 | 76.48 | 23.52 | 3.25 | 0.07 |
| phoneme | 5404 | 5 | 70.65 | 29.35 | 2.41 | 0.09 |
| pima | 768 | 8 | 65.10 | 34.90 | 1.87 | 0.09 |

Table A.2 Binary Balanced Datasets

| Data | #Instances | #Attributes | % of Class1 | % of Class2 | IR | SC |
|---|---|---|---|---|---|---|
| setosa_vs_versicolor | 100 | 4 | 50 | 50 | 1 | 0.6337 |
| setosa_vs_virginica | 100 | 4 | 50 | 50 | 1 | 0.6535 |
| virginica_vs_versicolor | 100 | 4 | 50 | 50 | 1 | 0.3316 |

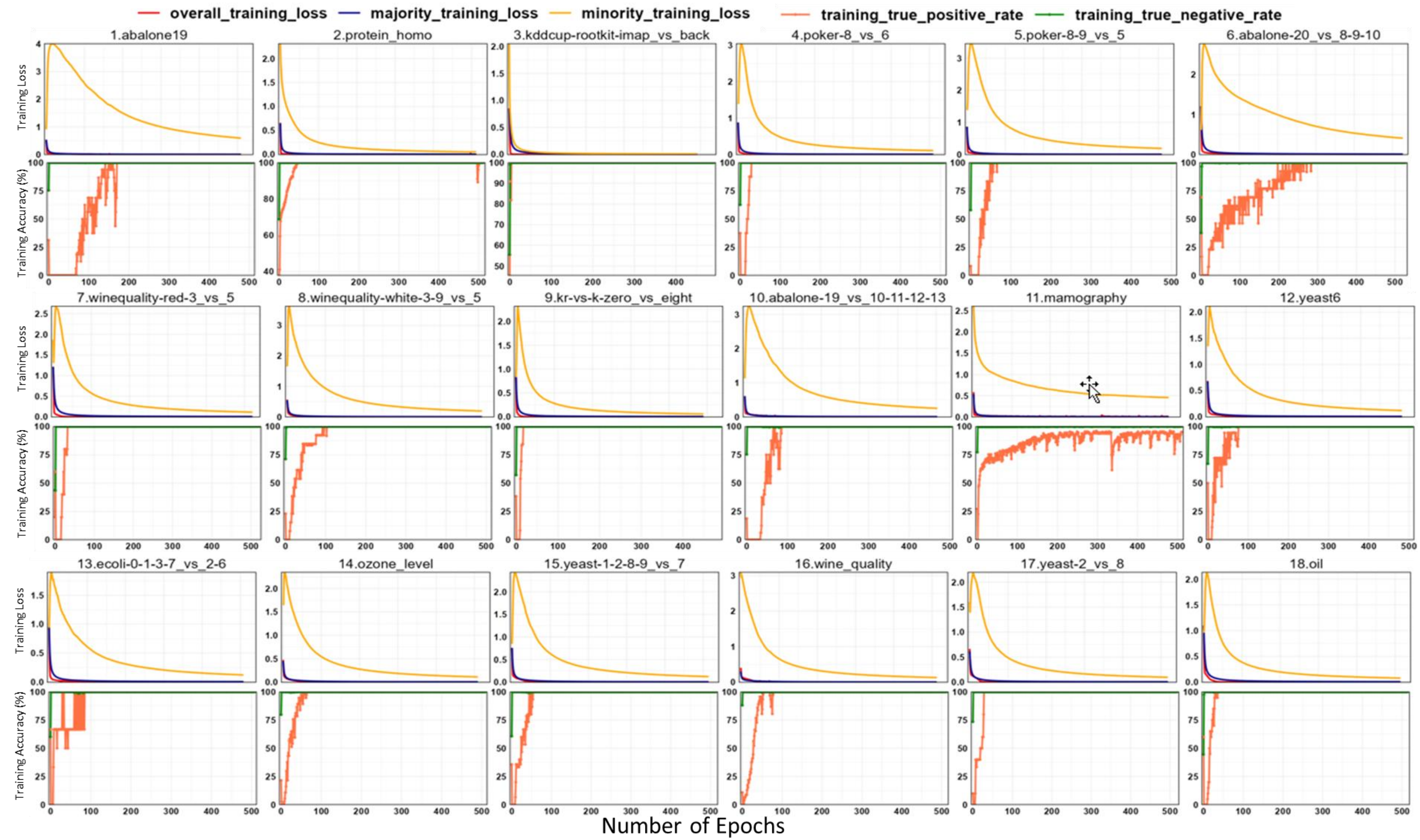

Fig A.1 Variation in Training Losses and Accuracies of Imbalanced Dataset Across Training Epochs

Fig A.1 Variation in Training Losses and Accuracies of Imbalanced Dataset Across Training Epochs (Continued)